\documentclass[a4paper]{article}

\usepackage{INTERSPEECH2022}

\usepackage{graphicx,subcaption}
\graphicspath{ {./images/} }

\title{Large-Scale Streaming End-to-End Speech Translation with Neural Transducers}
\name{Jian Xue*$^1$, Peidong Wang*$^1$, Jinyu Li$^1$, Matt Post$^2$, Yashesh Gaur$^1$ \thanks{* Equal Contribution}}
\address{
  $^1$Microsoft Speech Group\\
  $^2$Microsoft Translator Group}
\email{\{jiaxue, peidongwang, jinyli, mattpost, yagaur\}@microsoft.com}

\begin{document}

\maketitle
\begin{abstract}
Neural transducers have been widely used in automatic speech recognition (ASR). In this paper, we introduce it to streaming end-to-end speech translation (ST), which aims to convert audio signals to texts in other languages directly. Compared with cascaded ST that performs ASR followed by text-based machine translation (MT), the proposed Transformer transducer (TT)-based ST model drastically reduces inference latency, exploits speech information, and avoids error propagation from ASR to MT. To improve the modeling capacity, we propose attention pooling for the joint network in TT. In addition, we extend TT-based ST to multilingual ST, which generates texts of multiple languages at the same time. Experimental results on a large-scale 50 thousand (K) hours pseudo-labeled training set show that TT-based ST not only significantly reduces inference time but also outperforms non-streaming cascaded ST for English-German translation.

\end{abstract}

\section{Introduction}
\label{sec:intro}
Speech translation (ST) aims to convert speech signals to texts in other languages. Conventionally, it is formulated as a two-step cascaded task, automatic speech recognition (ASR) followed by text-based machine translation (MT) \cite{Ney1999ST, Matusov2005ST, Post2013ST}. Such cascaded systems typically suffer from the following issues. First, errors in ASR may propagate to MT. Second, since the intermediate representation is text, cascaded systems cannot fully leverage speech information (e.g., prosody) for translation. Finally, the MT module cannot start until the ASR module has (partially) finished, resulting in long inference latency.
Recently, end-to-end (E2E) ST (i.e., direct ST), which directly maps audio features to texts, has become more and more popular \cite{vila2018end,sperber2020speech}. In \cite{Berard2016ST}, the authors propose to use attention-based E2E encoder-decoder models (AED) \cite{chan2015listen,wang21t_interspeech} on a small French-English synthetic corpus. In \cite{weiss2017sequence}, a similar model structure is applied to the Fisher Callhome Spanish-English task and outperforms the cascaded method on the Fisher test set. AED-based models were also used in \cite{Berard2018ST} for a large-scale E2E ST task. However, AED models are usually operated in an offline mode which cannot start decoding until the full utterance is observed. 

E2E ST and ASR are similar in that they are both sequence-to-sequence mappings. Many model architectures can thus be shared, especially between ST using monotonic alignments \cite{raffel2017online} and ASR. To enable more effective communication between users, streaming (i.e., simultaneous) models are topics of investigation in both areas. Monotonic chunkwise attention (MoChA) \cite{chiu2018monotonic} was used in both MT and ASR. The MT version was extended to monotonic infinite lookback attention (MILk) \cite{arivazhagan2019monotonic} and monotonic multi-head attention \cite{ma2019monotonic, ma2021streaming}, and the ASR version was improved by multitask learning \cite{miao2019online} and minimum latency training strategies \cite{inaguma2020minimum}. Another streaming model architecture is the Neural transducer \cite{prabhavalkar2017asr,sainath2020asr,li2020asr,saon2021asr}, which outperforms MoChA and has emerged to be the state-of-the-art (SOTA) streaming E2E model in ASR \cite{li2021recent}, but has been less investigated in ST. Recently, Liu \emph{et al.} proposed cross attention augmented transducer (CAAT) for ST \cite{liu2021caat}. It uses Transformers in the joint network to combine encoder and prediction network outputs. Due to the use of Transformers and multi-step decision for memory footprint reduction, the latency of CAAT is large. In addition, to train a CAAT, complicated regularization terms and extensive hyper-parameter tuning are required.

In this paper, to leverage the success of the SOTA streaming technology in ASR, we propose to use neural transducers, specifically a low-latency and low-computational-cost Transformer transducer (TT) \cite{xiechen2021tt} for streaming E2E ST. To improve the representation fusion ability of the joint networks in TT, we propose \textit{attention pooling}. In addition, we extend TT for multilingual ST. The TT models are trained on a 50K-hour pseudo-labeled ST data set, which is generated by feeding the reference texts in ASR corpus to a MT model \cite{liu2019,jia2019st,gaido2020end}. We do not use any human-labeled paired ST data in this work. Experimental results on the Microsoft speech language translation (MSLT) corpus \cite{federmannmicrosoft} demonstrate that the proposed method not only achieves good bilingual evaluation understudy (BLEU) scores but also significantly reduces inference latency. The remainder of this paper is organized as follows. We describe the proposed methods and model structures in Section \ref{sec:sys}. The experimental setup and evaluation results are shown in Section \ref{sec:exp} and \ref{sec:eval}. We conclude this paper in Section \ref{sec:conclusion}.

\section{System Description}
\label{sec:sys}

\subsection{From Wait-$k$ To Neural Transducers}
\label{ssec:sys_transducer}

\begin{figure}[ht]
    \centering
    \begin{subfigure}[b]{.2\textwidth}
        \includegraphics[width=\linewidth]{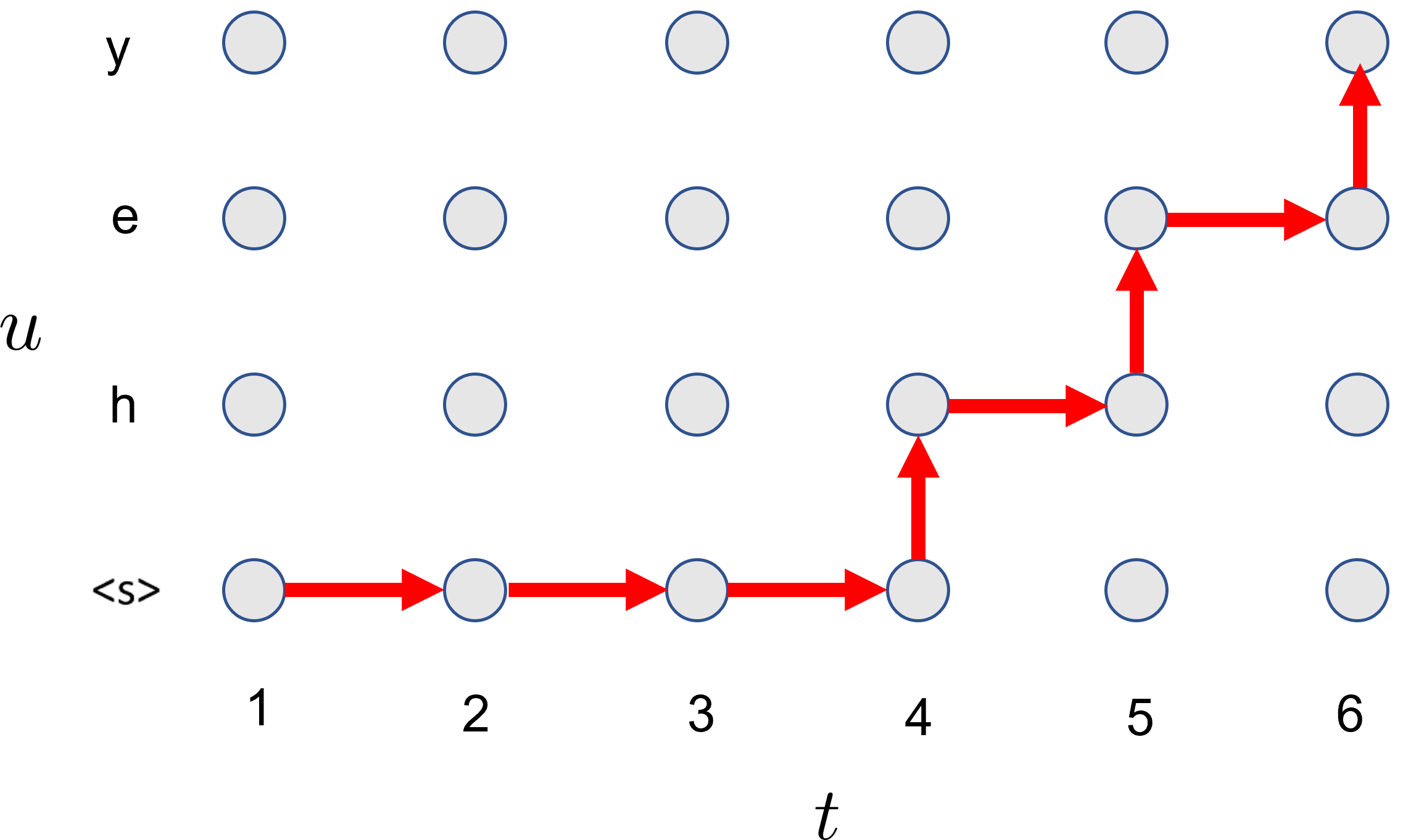}
        \caption{wait-3}
        \label{fig:wait-3}
    \end{subfigure}%
    \qquad
    \begin{subfigure}[b]{.2\textwidth}
        \includegraphics[width=\linewidth]{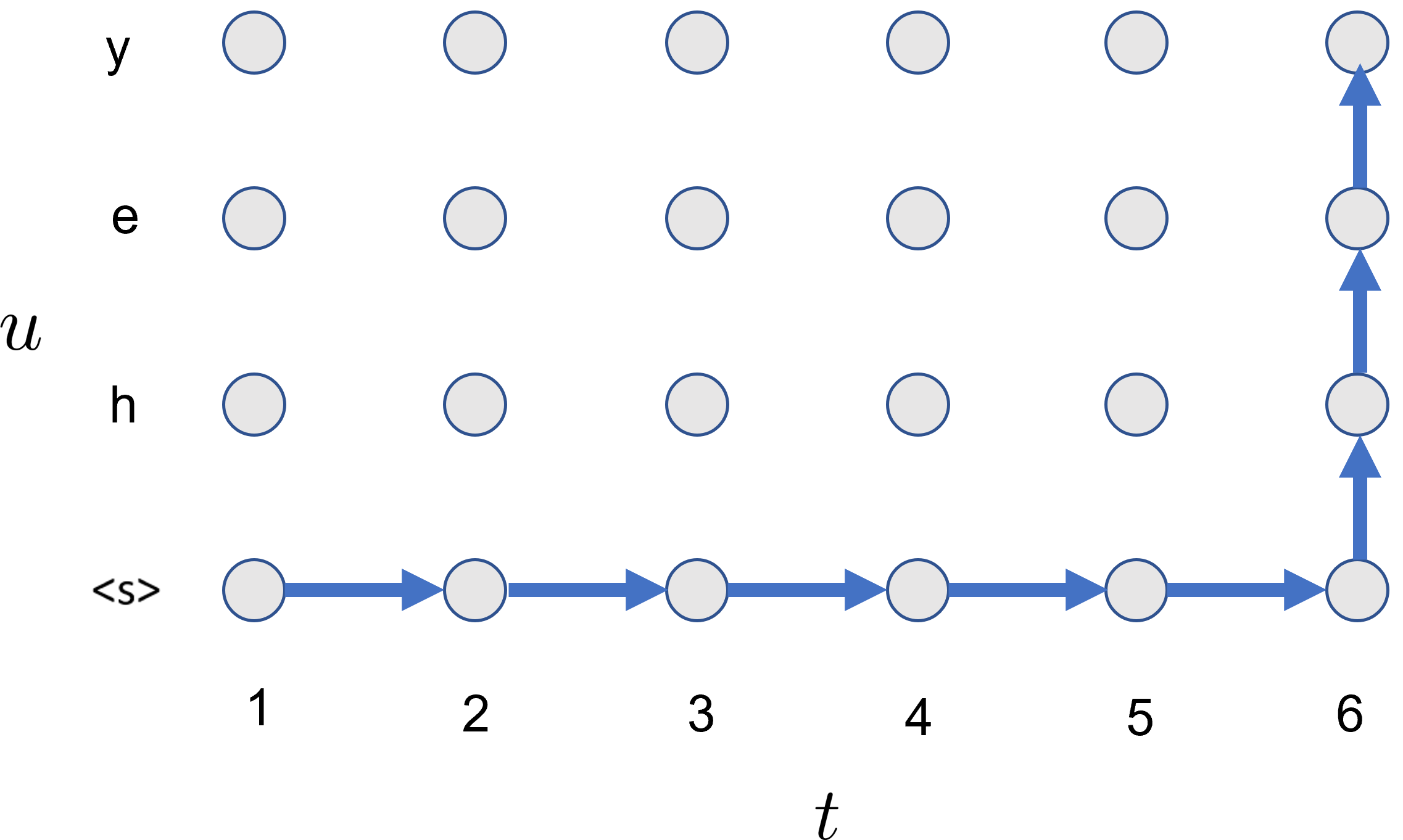}
        \caption{wait-$\infty$}
        \label{fig:wait-inf}
    \end{subfigure}%
    \centering
     \caption{Illustration of the decoding graphs of wait-$k$. For a given $k$, the read-write path is deterministic.}
    \label{fig:wait-k}
\end{figure}

Figure \ref{fig:wait-k} shows the decoding graphs of the commonly adopted wait-$k$ algorithm \cite{ma2019stacl} for streaming ST, where $t \in [1, T]$ and $u \in [1, U]$ denote the time steps for encoder output and output labels (i.e., read and write operations), respectively. As indicated by the names, wait-$3$ in Figure \ref{fig:wait-3} waits for 3 read operations to start writing, whereas wait-$\infty$ in Figure \ref{fig:wait-inf} can access the whole sentence.

\begin{figure}[ht]
    \centering
    \includegraphics[width=0.35\textwidth]{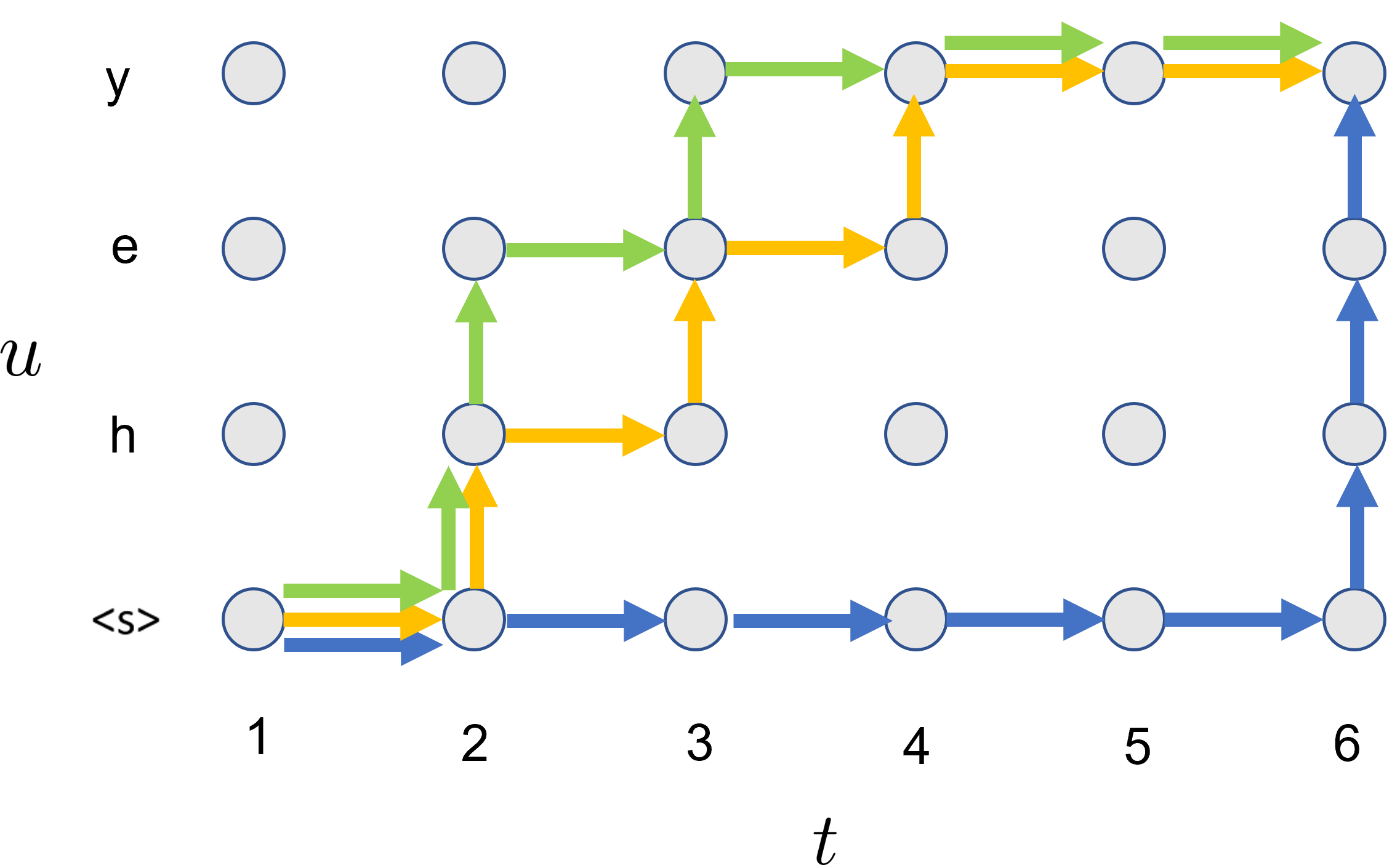}
    \caption{Illustration of three possible decoding graphs of a neural transducer. The path is chosen dynamically.}
    \label{fig:transducer_dec}
\end{figure}

Instead of using hard-coded wait steps and a fixed read-write policy in wait-$k$, neural transducers, whose decoding graphs are depicted in Figure \ref{fig:transducer_dec}, make read and write decisions in a data-driven fashion. 
During training, a neural transducer considers all possible alignments between encoder output and labels. At test time, it generates the most likely paths adaptively based on the input features. As shown in Figure \ref{fig:transducer_dec}, if there is no significant word reordering, the neural transducer may follow the orange path 
or a different green path. If there is a significant word reordering at the end of the utterance, it can use the blue decoding path corresponding to wait-$\infty$.

\begin{figure}[ht]
    \centering
    \includegraphics[width=0.35\textwidth]{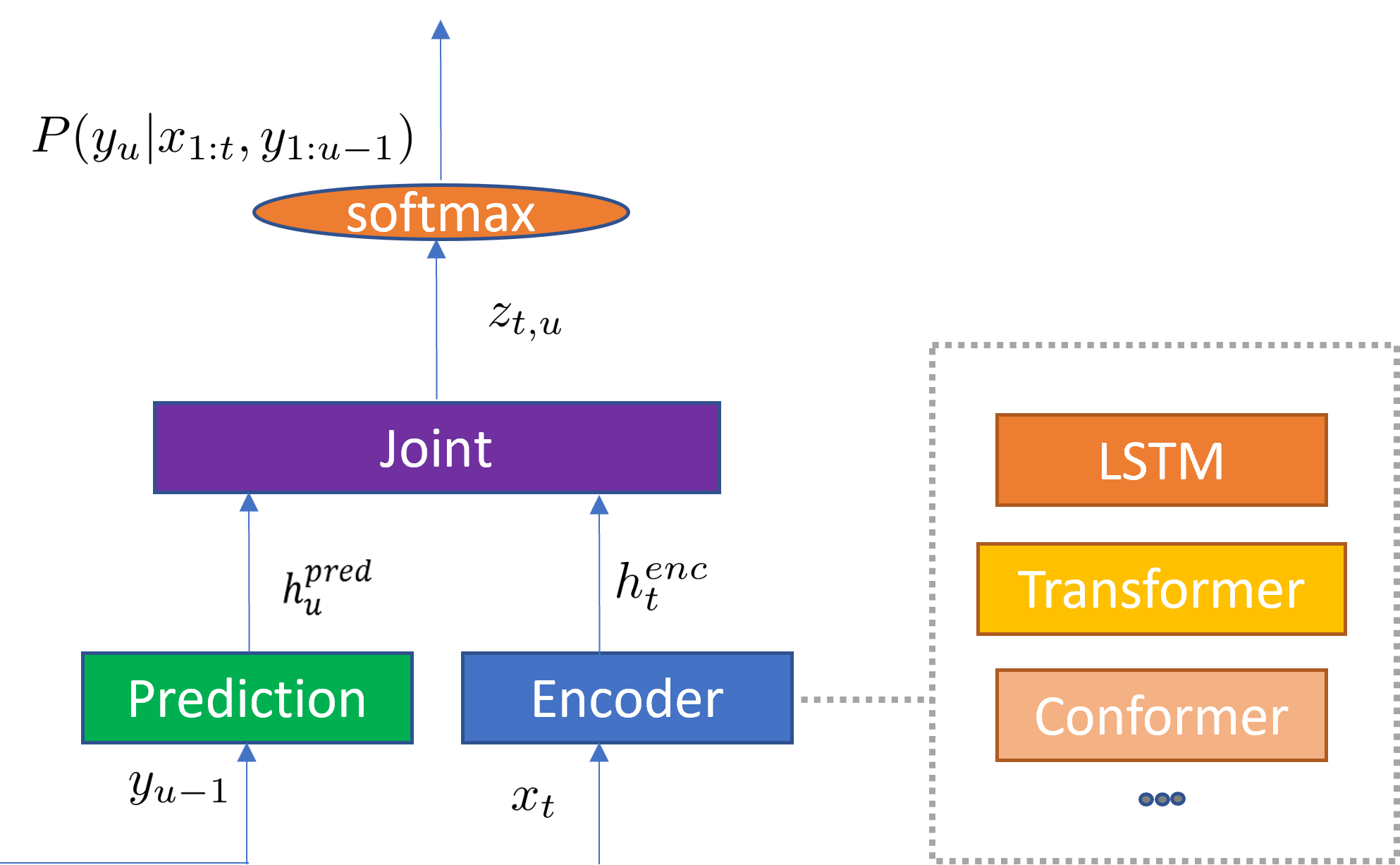}
    \caption{Illustration of neural transducers.}
    \label{fig:TT_fig}
\end{figure}

The model structure of neural transducers is shown in Figure \ref{fig:TT_fig}. It has three components: an encoder network, a prediction network, and a joint network. The encoder takes $d_x$-dimension audio features $\textbf{x}_t \in \mathbb{R}^{d_x}$ as input and generates $d_e$-dimension hidden representations $\textbf{h}_t^{\mathrm{enc}} \in \mathbb{R}^{d_e}$. The prediction network uses the embedding of non-blank output token $\textbf{y}_{u-1} \in \mathbb{R}^{1}$ at time $u-1$ and predicts hidden representation $\textbf{h}_u^{\mathrm{pred}} \in \mathbb{R}^{d_p}$ for step $u$. As for the joint network, it combines $\textbf{h}_t^{\mathrm{enc}}$ and $\textbf{h}_u^{\mathrm{pred}}$ to a $T \times U$ tensor, whose element at $t$ and $u$ is denoted by the vector $\textbf{z}_{t,u} \in \mathbb{R}^{d_z}$. After a softmax operation, the model generates probability $P(\textbf{y}_u \in \textbf{Y} \cup \varnothing |\textbf{x}_{1:t}, \textbf{y}_{1:u-1})$, where $\textbf{Y}$ is the vocabulary list and $\varnothing$ denotes the \textit{blank} output (i.e., output nothing). Note that for notation simplicity, we ignore batch size and use the same time resolution for $\textbf{x}_t$ and $\textbf{h}_t^{\mathrm{enc}}$ in this paper. Neural transducers can use different types of networks as encoders, such as long short-term memory (LSTM) recurrent neural networks (RNNs) in RNN-T models \cite{graves2012asr} and Transformers in TT models \cite{facebook2019TT, zhang2020tt}.

\subsection{Streaming TT Model}
\label{ssec:streaming_TT}
\begin{figure}[ht]
    \centering
    \includegraphics[width=0.35\textwidth]{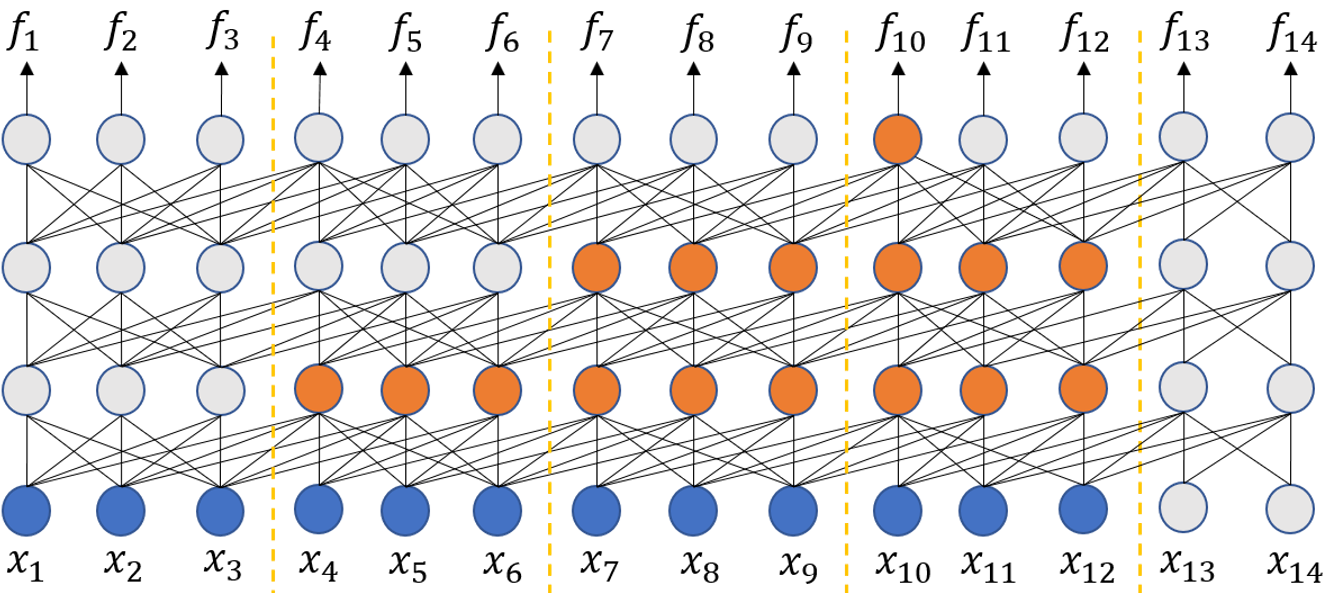}
    \caption{Illustration of the reception field of a streaming TT at position $f_{10}$ with chunk size 3 and the number of left chunks 1.}
    \label{fig:reception_fig}
\end{figure}

We apply TT in this work since it usually obtains better performance than RNN-T in ASR tasks \cite{xiechen2021tt, facebook2019TT, zhang2020tt}. Each Transformer block in the encoder is constructed from a multi-head self-attention layer followed by a feedforward layer. In order for TT to work in a streaming mode with low latency and low computational cost, we apply the attention mask proposed in \cite{xiechen2021tt}. The attention mask can be the same for different layers. At each layer $l$, we divide the input $\textbf{x}^l_{1:T}$ into chunks $\textbf{c}^l_{1:S}$ along time with chunk size $U$, where $\textbf{c}^l_{s} = \textbf{x}^l_{s \times U: (s+1) \times U - 1}$. At time step $t$, $\textbf{x}^l_t$ can only see the frames inside its own chunk $\textbf{c}^l_{t / U + 1}$ and a fixed number $B$ of left chunks $\textbf{c}^l_{max(1, t / U + 1 - B):t / U}$. Figure \ref{fig:reception_fig} shows an example of the reception field for a three-layer Transformer model with chunk size $U=3$ and the number of left chunks $B=1$ at output position $f_{10}$. Note that since the features cannot access the frames ahead its own chunk, as shown by the right-most blue circle in the first input layer, the number of look-ahead frames is kept $U-1=2$. Moreover, the left reception field increases linearly with the number of layers, enabling the model to use a long history information for a better performance.

\subsection{Attention Pooling for Joint Networks}
\label{ssec:sys_pool}
The joint network in a conventional neural transducer combines the output representations of encoder and prediction network with simple linear layers:
\begin{equation}
    \textbf{z}_{t,u} = W^{\mathrm{out}} f(W_e^{\mathrm{joint}} \textbf{h}_t^{\mathrm{enc}} + W_p^{\mathrm{joint}} \textbf{h}_u^{\mathrm{pred}})
    \label{eq:conv}
\end{equation}
where the two sources of output are multiplied with $W_e^{\mathrm{joint}} \in \mathbb{R}^{d_e \times d_j}$ and $W_p^{\mathrm{joint}} \in \mathbb{R}^{d_p \times d_j}$, to map the feature vectors to $d_j$-dimension, respectively. $f$ denotes a non-linear function, which is typically $\tanh$ or $relu$. Finally, the feature vector is converted to the output dimension using $W^{\mathrm{out}} \in \mathbb{R}^{d_j \times d_z}$. 

Recent study in ASR shows that the representation fusion ability of such joint networks can be improved by a bilinear pooling approach \cite{zhang2022improving}. In this paper, we propose attention pooling, which could {adapt the pooling weights according to the input using an attention-like weighting mechanism}.
Different from ASR, ST needs to consider not only the current output probability $P(\textbf{y}_u \in \textbf{Y} \cup \varnothing |\textbf{x}_{1:t}, \textbf{y}_{1:u-1})$ but also whether writing a non-\textit{blank} token at a future step is better. The adaptive attention weights in attention pooling may work as an additional type of feature to help ST models to make more appropriate decisions. Note that we keep the time and space complexity of attention pooling to be linear so that it consumes less computation resources during inference. The proposed attention pooling is defined as Equation (\ref{eq:att_out}) to (\ref{eq:att_dec}) below:
\begin{equation}
    \textbf{z}_{t,u} = W^{\mathrm{out}} f(\hat{\textbf{h}}_{t,u}^{\mathrm{joint}} + W_e^{\mathrm{joint}} \textbf{h}_t^{\mathrm{enc}} + W_p^{\mathrm{joint}} \textbf{h}_u^{\mathrm{pred}})
    \label{eq:att_out}
\end{equation}
\begin{equation}
    \hat{\textbf{h}}_{t,u}^{\mathrm{joint}} = W^{\mathrm{proj}} ({v}_t^{\mathrm{enc}} \odot {v}_u^{\mathrm{pred}})
    \label{eq:att_intermediate}
\end{equation}
\begin{equation}
    {v}_t^{\mathrm{enc}} = f((softmax(W^e \textbf{h}_t^{\mathrm{enc}}) \cdot \textbf{h}_t^{\mathrm{enc}}))
    \label{eq:att_enc}
\end{equation}
\begin{equation}
    {v}_u^{\mathrm{pred}} = f((softmax(W^p \textbf{h}_u^{\mathrm{pred}}) \cdot \textbf{h}_u^{\mathrm{pred}}))
    \label{eq:att_dec}
\end{equation}
where $\hat{\textbf{h}}_{t,u}^{\mathrm{joint}} \in \mathbb{R}^{d_j}$ is the pooling term at time steps $t$ and $u$, $W^{\mathrm{proj}} \in \mathbb{R}^{1 \times d_j}$ maps the 1-dimension feature to $d_j$-dimension, ${v}_t^{\mathrm{enc}} \in \mathbb{R}^{1}$ and ${v}_u^{\mathrm{pred}} \in \mathbb{R}^{1}$ denotes the contribution of encoder and prediction network to the pooling term, $\odot$ denotes Hadamard product, $W^e \in \mathbb{R}^{d_e \times d_e}$ and $W^p \in \mathbb{R}^{d_p \times d_p}$ are used to calculate the attention weights, and $\cdot$ denotes tensor-dot operation.

We also design a stronger qkv attention pooling method that uses separate weights for query, key, and value. It can be expressed by replacing Equation (\ref{eq:att_enc}) and (\ref{eq:att_dec}) with Equation (\ref{eq:att_qkv_enc}) and (\ref{eq:att_qkv_dec}), respectively:

\begin{equation}
    {v}_t^{\mathrm{enc}} = f((softmax(W^e_q \textbf{h}_t^{\mathrm{enc}} \odot W^e_k \textbf{h}_t^{\mathrm{enc}}) \cdot (W^e_v \textbf{h}_t^{\mathrm{enc}}))
    \label{eq:att_qkv_enc}
\end{equation}
\begin{equation}
    {v}_u^{\mathrm{pred}} = f((softmax(W^p_q \textbf{h}_u^{\mathrm{pred}} \odot W^p_k \textbf{h}_u^{\mathrm{pred}}) \cdot (W^p_v \textbf{h}_u^{\mathrm{pred}})))
    \label{eq:att_qkv_dec}
\end{equation}
where $W^e_q \in \mathbb{R}^{d_e \times d_e}$, $W^e_k \in \mathbb{R}^{d_e \times d_e}$, and $W^e_v \in \mathbb{R}^{d_e \times d_e}$ are the linear layers for query, key, and value for encoder features, and $W^p_q \in \mathbb{R}^{d_p \times d_p}$, $W^p_k \in \mathbb{R}^{d_p \times d_p}$, and $W^p_v \in \mathbb{R}^{d_p \times d_p}$ are the corresponding weights for the prediction network. Note that we use Hadamard product and tensor-dot to avoid quadratic time and space complexity.

\subsection{Multilingual ST with TT}
\label{ssec:sys_multilingual}
ST supporting a single language pair such as English-Chinese (EN-ZH) are often referred to as bilingual ST. It is inefficient to build a separate bilingual ST model for every language pair. In addition, running multiple bilingual ST models simultaneously requires a lot of memory and computation resources. In this work, we propose to apply TT to multilingual ST by sharing the encoder and using separate prediction and joint networks for different target languages. Since the encoder (64M of parameters in our experiments) is much larger than joint and prediction networks (24M combined), the size of such a multilingual ST model is comparable to a bilingual model.

\begin{figure}[ht]
    \centering
    \includegraphics[width=0.35\textwidth]{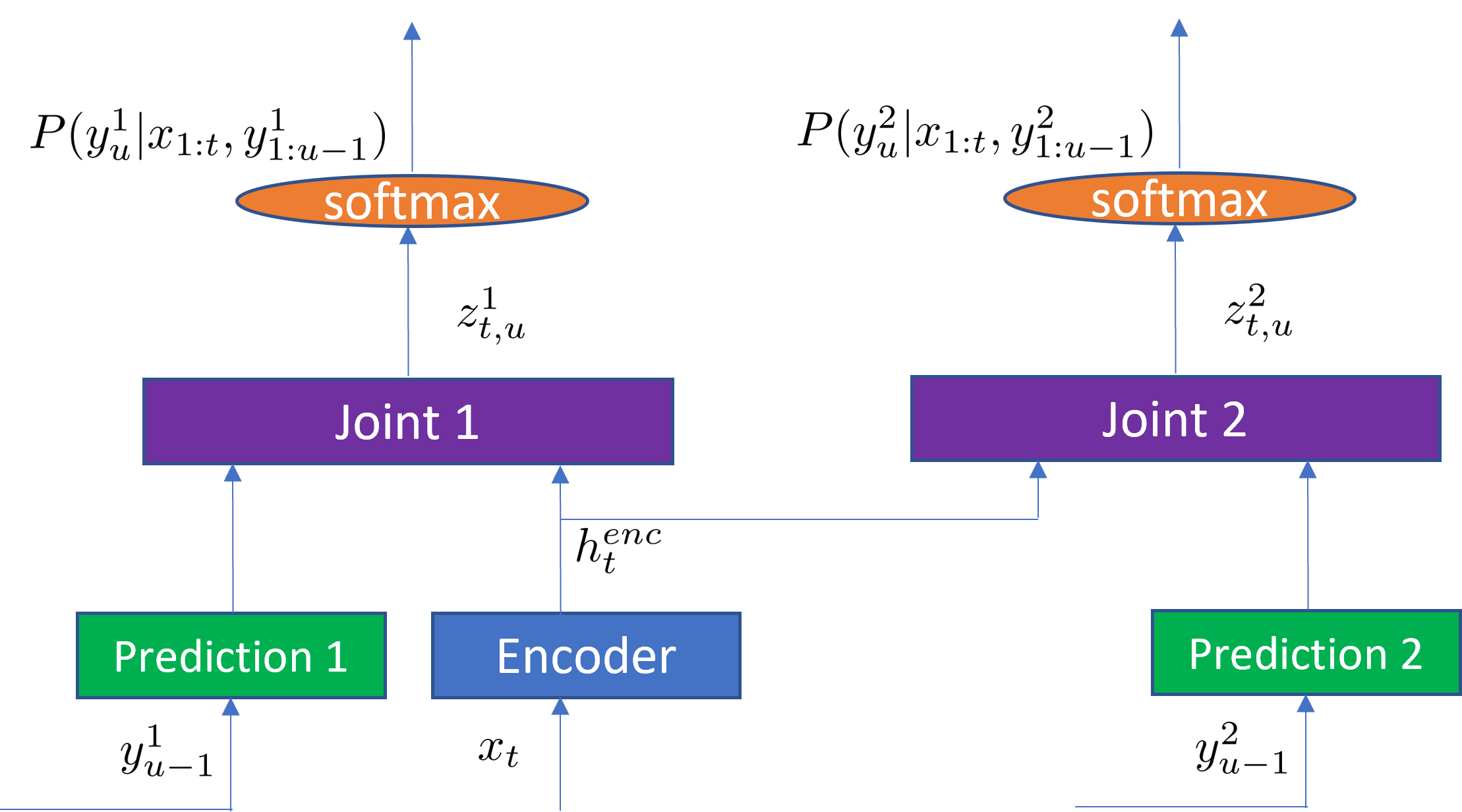}
    \caption{Illustration of TT for multilingual ST.}
    \label{fig:multilingual_fig}
\end{figure}

Figure \ref{fig:multilingual_fig} shows a one-to-many multilingual ST model using TT. Shared encoder output $\textbf{h}_t^{\mathrm{enc}}$ is fed to multiple prediction and joint networks. During training, we alternate the training data for different batches, e.g., one batch using EN-ZH data and another English-German (EN-DE) (i.e., data mixing ratio is 50\%-50\%). 

\section{Experimental Setup}
\label{sec:exp}
We use 50K hours of Microsoft internal data to train ASR and E2E ST models. The EN-ZH MT model is trained with 62M lines of parallel data plus 110M lines of back-translated mono-lingual data, and the EN-DE MT model uses 162M lines of parallel data plus 110M lines of back-translated mono-lingual data. All the data are anonymized with personally identifiable information removed. The original transcriptions are in English, and we use Microsoft cognitive translation service to translate them into Chinese and German. We do not use any human-labeled paired ST data. For evaluation, we use the publicly available MSLT\_v1.0 for EN-DE and MSLT\_v1.1 for EN-ZH \cite{federmannmicrosoft}. \footnote{We are not allowed to use other public data sets due to license restrictions.} 

\subsection{Cascaded Method}
\label{sssec:cascaded}
We use the cascaded method as the baseline for our experiments.
The ASR module is a streaming TT model described in Section \ref{ssec:streaming_TT}. It is trained using the above 50K-hour English audio and the corresponding English transcriptions. The encoder consists of 18 Transformer blocks, each containing 320 hidden nodes, 8 attention heads, and 2048 feedforward nodes. The prediction network uses 2 LSTM layers. Each LSTM layer has 1024 hidden nodes and the embedding dimension is also 1024. The joint network is a simple feedforward layer of size 512. The vocabulary size is 4K and the model has 88M parameters. 
The model input is 80-dimension log-Mel filter-bank features with 25ms windows and 10ms shift, extracted from mixed band training data \cite{li2012improving}. 
Before the input is fed to the Transformer blocks, it is filtered and down-sampled by a factor of 4 using two convolutional layers.
The chunk size of the streaming mask for the Transformer blocks is 4.
The output of the ASR module is used by a non-streaming MT module to get the translation results. 
The MT model consists of a 6-layer Transformer encoder and a 2-layer RNN decoder. Each Transformer layer has a feedforward network of size 2048 and 8 attention heads. The embedding size is 512. The MT model has 67M parameters.

\subsection{Streaming E2E ST Models}
\label{sssec:transducer}
The streaming E2E ST models are trained using the same 50K-hour English audio, but with the corresponding translated labels generated by the MT model.
The ST models have the same architecture as that of the ASR module in the above cascaded system, except that the output dimensions are changed according to vocabulary sizes. The vocabulary size of EN-ZH is 11K, whereas that of EN-DE is 4K. In addition to chunk size 4, which has 160ms look-ahead and is denoted as TT-160ms, we conduct experiments using a chunk size of 80, whose look-ahead is 3.2s and is thus denoted as TT-3.2s. 

\subsection{ASR Encoder Initialization and ASR Multi-Task Learning} 
\label{sssec:init_multi}
In addition to pseudo-labeling, we investigate ASR encoder initialization and ASR multi-task learning for ST.
For ASR encoder initialization, we use the above ASR model to initialize the encoder and randomly initialize the prediction and joint networks. Then we fine-tune the whole model. 
For ASR multi-task learning, we adopt the multilingual ST model described in Section \ref{ssec:sys_multilingual}, but with English and Chinese as the output languages. 

\subsection{Attention Pooling for Joint Networks}
\label{sssec:att}
Each new weight matrix introduced by attention pooling is implemented as a linear layer. 
The open neural network exchange (ONNX) conversion procedures are modified accordingly.


\subsection{ONNX Conversion and Model Compression}
\label{sssec:onnx}
After getting the checkpoints of TT models, we convert them to ONNX format, compress each component, and evaluate the compressed models. The weights in the encoders are compressed to uint8, and the RNN and feedforward layers in the prediction and joint networks are compressed using neural-network unified preprocessing heterogeneous architecture (NUPHAR). Quantization does not cause significant accuracy degradation.


\subsection{Latency Measurement}
\label{sssec:latency}
We use average proportion (AP), average lagging (AL), and differentiable average lagging (DAL) proposed in \cite{ma2020simuleval} to measure the inference latencies of our ST systems. Note that different from \cite{ma2020simuleval}, our results are generated after the models are converted to ONNX format and compressed.


\section{Evaluation Results}
\label{sec:eval}

\subsection{Transformer Transducer for Speech Translation}
\label{ssec:res_transducer}
The first parts of Table \ref{tab:results_zh} and \ref{tab:results_de} contain the BLEU scores and latencies of the cascaded and E2E ST models. Note that no external language model is used for both methods, and the average length of evaluation utterances is 4.4s \cite{federmannmicrosoft}.
First, TT-3.2s outperforms TT-160ms in BLEU scores on both EN-ZH and EN-DE, at the cost of a significant latency increase. The differences between TT-3.2s and TT-160ms in BLEU scores are small: 0.7 on EN-ZH and 1.3 on EN-DE, indicating that TT can maintain a good translation quality when working in streaming mode.
The AL of TT-3.2s is about 2.2s, shorter than the look-ahead time. The reason is that except for the first output token, TT-3.2s does not have to wait for the whole 3.2s to generate an output. On the contrary, the AL of TT-160ms is 841ms, longer than 160ms. This shows that that TT-160ms requires multiple frames to handle word reordering.
Second, comparing the BLEU scores of cascaded models and TT-160ms, we observe that on EN-ZH, there is still a gap. However, on the EN-DE test set, TT-160ms outperforms the cascaded model. Note that TT-160ms is a streaming model with a very small latency, whereas the cascaded model is a non-streaming model and is trained using additional text-text MT data. 
Finally, we mainly use pseudo-labeling to deal with the data scarcity problem in this study. To exploit the 50K-hour ASR training data, we investigate ASR encoder initialization and ASR multi-task learning as described in Section \ref{sssec:init_multi}. As shown in Table \ref{tab:results_zh}, these two methods do not help in our experiments, possibly because our ST models are trained with a large amount of training data. 

\subsection{Attention Pooling for Joint Networks}
\label{ssec:res_attetion_pooling}
Table \ref{tab:results_zh} contains the comparison between TT-160ms and different pooling methods for joint networks. 
Bilinear pooling does not improve the performance of TT-160ms in this study. The reason may be that it lacks the ability to adapt the pooling weights according to the input, which is important in ST since the output of the prediction network is in a different language and is not monotonic w.r.t. the audio features. 
The attention pooling methods proposed in this paper show consistent BLEU score improvements over TT-160ms. The latencies are also very close to those of TT-160ms. Note that each input frame is 10ms and the encoder has a subsampling factor of 4. The attention pooling methods are thus at most 1-2 steps slower at the encoder output, and the slightly higher latency of simple attention pooling over qkv attention pooling can be negligible.
Since the simple attention pooling method obtains a larger BLEU score improvement per additional parameter, which is calculated as $\Delta BLEU / \Delta \#params$, we use it for the evaluation on EN-DE. As shown in Table \ref{tab:results_de}, attention pooling obtains a consistent BLEU score improvement over TT-160ms.

\subsection{Multilingual ST with TT}
\label{ssec:res_multilingual_ST}
The last lines in Table \ref{tab:results_zh} and Table \ref{tab:results_de} correspond to the EN-ZH output and EN-DE output of the TT-based streaming E2E multilingual ST. Note that the multilingual ST results in the above tables do not use attention pooling, and the EN-ZH and EN-DE outputs are generated simultaneously. The BLEU scores of multilingual ST are close to those of the bilingual TT-160ms models.
In addition, multilingual ST greatly reduces the model size and computation burden since it shares a single encoder for multiple languages.

%

\begin{table}[t]
  \caption{Comparisons of BLEU scores and latencies on EN-ZH of MSLT\_v1.1\_test. The numbers following the pooling methods denote the number of additional parameters being introduced. The AL and DAL values are in milliseconds (ms).} 
  \label{tab:results_zh}
  \setlength{\tabcolsep}{3pt}
  \centering
  \begin{tabular}{l c c c c c c}
    \toprule
    methods & BLEU $\uparrow$ & AP $\downarrow$ & AL $\downarrow$ & DAL $\downarrow$ \\
    \midrule
    cascaded & 40.0 & 1 & $\infty$ & $\infty$ \\
    TT-3.2s & 35.6 & 0.74 & 2151  & 1886 \\
    TT-160ms & 34.9 & 0.61 & 841 & 834 \\
    \hspace{3mm}ASR encoder init & 34.7 & 0.61 & 841 & 834 \\
    \hspace{3mm}ASR multi-task learning & 34.7 & 0.61 & 841 & 834 \\
    \hspace{3mm}bilinear (+2K) & 34.5 & 0.61 & 862 & 859 \\
    \hspace{3mm}attention (+1.3M) & 35.1 & 0.62 & 910 & 910 \\
    \hspace{3mm}qkv attention (+3.9M) & {35.3} & 0.62 & 875 & 877 \\
    \midrule
    multilingual EN-ZH output & 34.8 & 0.61 & 841 & 834 \\
    \bottomrule
  \end{tabular}
\end{table}

\begin{table}[t]
  \caption{Comparisons of BLEU scores and latencies on EN-DE of MSLT\_v1.0\_test.}
  \label{tab:results_de}
  \setlength{\tabcolsep}{3pt}
  \centering
  \begin{tabular}{l c c c c c }
    \toprule
    methods & BLEU $\uparrow$ & AP $\downarrow$ & AL $\downarrow$ & DAL $\downarrow$ \\
    \midrule
    cascaded & 29.3 & 1 & $\infty$ & $\infty$ \\
    TT-3.2s & 30.7 & 0.74 & 2152 &  1890 \\
    TT-160ms & 29.4 & 0.61 & 828 & 828 \\
    \hspace{3mm}attention (+1.3M)  & 29.6 & 0.61 & 871 & 869 \\ 
    \midrule
    multilingual EN-DE output & 29.2 & 0.61 & 828 & 828 \\ %
    \bottomrule
  \end{tabular}
\end{table}


  

  

\section{Conclusions}
\label{sec:conclusion}
We propose neural transducers for large-scale streaming E2E ST. To improve the performance of TT for ST, we propose attention pooling for joint networks. Moreover, we extend TT to multilingual ST by sharing the encoder. Experimental results on the EN-ZH and EN-DE test sets of MSLT show that the proposed TT-based streaming E2E ST models achieve high-quality translation performance with low inference latency. More specifically, the proposed streaming E2E ST system outperforms a non-streaming cascaded system on EN-DE.
\section{Acknowledgement}
We would like to thank Drs. Long Zhou, Yu Wu, and Shujie Liu at Microsoft Research Asia for valuable suggestions.

\bibliographystyle{IEEEtran}

\bibliography{mybib}


\end{document}